\documentstyle{article}

\input{epsf}

\begin{document}

\title{Linear and Order Statistics Combiners for Pattern Classification
\footnote{Appears in {\em Combining Artificial Neural Networks}, Ed.
Amanda Sharkey, pp 127-162, Springer Verlag, 1999.}}

\author{Kagan Tumer\\
NASA Ames Research Center\\
Caelum Research \\
{\tt kagan@ptolemy.arc.nasa.gov} \\
\and
Joydeep Ghosh \\ 
Dept. of Elec. and Com Engr. \\ 
The University of Texass \\
{\tt ghosh@pine.ece.utexas.edu} \\
}

\date{}

\maketitle

\begin{abstract}
Several researchers have experimentally shown that substantial
improvements can be obtained in difficult pattern recognition
problems by combining or integrating the outputs of multiple 
classifiers.
This chapter provides an analytical framework to {\em quantify} the
improvements in classification results due to combining. The
results apply to both linear combiners and order statistics combiners.
We first show that to a first order approximation,
the error rate obtained over and above the Bayes error rate, is
directly proportional to the variance of the actual decision boundaries
around the Bayes optimum boundary. 
Combining classifiers in output space
reduces this variance, and hence reduces the "added" error.
If $N$ unbiased classifiers are combined by simple averaging, 
the added error rate can be reduced by a factor of $N$ if the 
individual errors in approximating the decision boundaries are uncorrelated.
Expressions are then derived for linear combiners which are biased
or correlated, and the effect of output correlations on
ensemble performance is quantified.
For order statistics based non-linear combiners, 
we derive expressions that indicate how much 
the median, the maximum and in general the $i$th order statistic 
can improve classifier performance.
The analysis presented here facilitates the
understanding of the relationships among error rates,
classifier boundary distributions, and combining in output space.
Experimental results on several public domain data sets
are provided to illustrate the benefits of combining and to support
the analytical results.
\end{abstract}

%\newpage

\section{Introduction}
\label{sec:intro}
Training a parametric classifier involves the use of a
{\em training} set of data with known labeling to estimate or 
``learn'' the parameters of the chosen model. 
A {\em test} set, consisting of patterns not previously seen by the 
classifier, is then used to determine the classification performance.
This ability to meaningfully respond to novel patterns, or generalize, 
is an important aspect of a
classifier system and in essence, the true gauge of
performance~\cite{leti90,wolp90b}. 
Given infinite training data, consistent classifiers approximate the
Bayesian decision boundaries to arbitrary precision, therefore
providing similar generalizations~\cite{gebi92}.
However, often only a limited portion of the pattern space is
available or observable~\cite{duha73,fuku90}. 
Given a finite and noisy data set, different 
classifiers typically provide different generalizations by realizing
different decision boundaries~\cite{ghtu94a}. 
For example, when classification is performed using a multilayered, 
feed-forward artificial neural network,
different weight initializations, or different architectures (number of
hidden units, hidden layers, node activation functions etc.) 
result in differences in 
performance. It is therefore beneficial to train an ensemble of
classifiers when approaching a classification problem to ensure that
a good model/parameter set is found.

\begin{figure}[htb]
\epsfxsize=4.9in \epsfbox{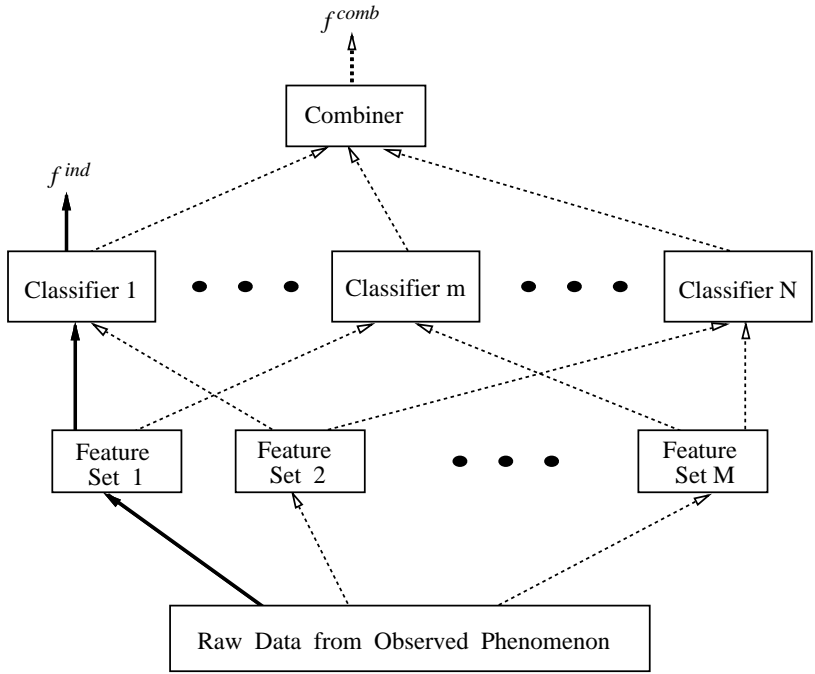}
  \caption{Combining Strategy. The solid lines leading to $f^{ind}$ 
	   represent the decision of a specific classifier, while the 
	   dashed lines lead to $f^{comb}$, the output of the combiner.}
  \label{fig:comb}
\end{figure}

Techniques such as cross-validation also lead to multiple trained
classifiers. Selecting
the ``best'' classifier is not necessarily
the ideal choice, since potentially valuable information may be
wasted by discarding the results of less-successful 
classifiers. This observation motivates the concept of ``combining'' wherein
the outputs of all the available
classifiers are pooled before a decision is made. This approach 
is particularly useful for difficult problems, such as those
that involve a large amount of noise, limited number of training
data, or unusually high dimensional patterns.
The concept of combining appeared in  the neural network literature
as early as 1965 \cite{nils65}, and has subsequently been studied in
several forms, including {\em stacking}~\cite{wolp92}, 
{\em boosting}~\cite{drsc93,drco94,frsc95,frsc96} 
and {\em bagging}~\cite{brei93,brei94}.
Combining has also been studied in other fields such as econometrics,
under the name ``forecast combining'' \cite{gran89}, or machine
learning where it is called ``evidence combination'' \cite{barn81,galo81}. 
The overall architecture of the combiner form studied in this
article is shown in Figure~\ref{fig:comb}.
The output of an individual classifier using a single feature set is
given by $f^{ind}$. Multiple classifiers, possibly trained on different
feature sets, provide the combined output $f^{comb}$.

Currently, the most popular way of combining multiple 
classifiers is via simple averaging of the corresponding
output values \cite{hawa92,lisk90,peco93,tugh96}.
Weighted averaging  has also been proposed, along with different methods 
of computing the proper classifier weights \cite{besv94,hasc93,lisk90,mepa97}. 
Such linear combining techniques have been mathematically analyzed for
both regression \cite{hasc93,peco93b} and classification~\cite{tugh96} 
problems.
Order statistics combiners that selectively pick a classifier on 
a per sample basis were introduced in~\cite{tugh95f,tugh98a}.
Other non-linear methods, such as  
rank-based combiners~\cite{alku95,hohu94}, or voting 
schemes~\cite{baco94,chst95,hasa90}
have been investigated as well. 
Methods for combining beliefs in the Dempster-Shafer sense
are also available \cite{rikn91,rogo94,xukr92,yasi94}.
Combiners have been successfully applied a multitude of real 
world problems \cite{baxt92,bigi88,ghtu96,lehw91,shli90,zhme92}.

Combining techniques such as 
majority voting can generally be applied to any type of 
classifier, while others rely on specific outputs, or specific 
interpretations of the output. For example, the confidence factors 
method found in machine learning literature relies on the interpretation of 
the outputs as the belief 
that a pattern belongs to a given class \cite{heck86}. 
The rationale for averaging, on the other hand, is based on the result that 
the outputs of parametric classifiers that are trained to minimize
a cross-entropy or mean square error (MSE) function, given
``{\em one-of-L}'' desired output patterns, approximate the {\em a posteriori} 
probability densities of the corresponding class \cite{rili91,ruro90}. 
In particular, the MSE is shown to be equivalent to:
\begin{eqnarray*} 
MSE = K_1 + \sum_i \int_x D_i(x)
\left(p(C_i|x) - f_i(x)\right)^2 dx
\end{eqnarray*} 
where $K_1$ and $D_i(x)$ depend on the class distributions only, $f_i(x)$ is
the output of the node representing class $i$ given an output $x$, 
$p(C_i|x)$ denotes the posterior probability and the summation is over 
all classes \cite{shca91}. Thus minimizing the MSE is equivalent to a weighted
least squares fit of the network outputs to the corresponding posterior 
probabilities.

In this article we first analytically study the effect of linear and
order statistics combining in 
output space with a focus on the relationship between decision
boundary distributions and error rates. 
Our objective is to provide an analysis that: 
\begin{itemize}
\item{} encapsulates the most commonly used combining strategy, averaging 
in output space;
\item{} is broad enough in scope to cover non-linear combiners; and
\item{} relates the location of the decision boundary to the classifier error.
\end{itemize}

The rest of this article is organized as follows. 
Section~\ref{sec:sing} introduces the overall framework for estimating
error rates and the effects of combining. 
In Section~\ref{sec:lin} we analyze linear combiners, and derive expressions
for the error rates for both biased and unbiased classifiers. 
In Section~\ref{sec:order}, we examine order statistics 
combiners, and analyze the resulting classifier boundaries and error regions.
In Section~\ref{sec:indep} we study linear combiners that make
correlated errors, derive their error reduction rates, and discuss
how to use this information to build better combiners.
In Section~\ref{sec:resu}, we present experimental results
based on real world problems, and we conclude with a discussion
of the implications of the work presented in this article. 

\section{Class Boundary Analysis and Error Regions}
\label{sec:sing}
Consider a single classifier whose outputs are
expected to approximate the corresponding {\em a posteriori}
class probabilities if it is reasonably well trained. 
The decision boundaries obtained by such a classifier are 
thus expected to be close to Bayesian decision boundaries. 
Moreover, these boundaries will tend to occur in regions where 
the number of training samples belonging to the two most locally dominant
classes (say, classes $i$ and $j$) are comparable.  

We will focus our analysis on network performance around the decision 
boundaries. Consider the boundary between classes $i$ and $j$ for a
single-dimensional input (the extension to multi-dimensional inputs
is discussed in ~\cite{tume96}). 
First, let us express the output response of the $i$th unit of a
{\em one-of-L} classifier network to a given input $x$ as\footnote{If 
two or more classifiers need to be distinguished, a superscript is 
added to $f_i(x)$ and $\epsilon_i(x)$ to indicate the classifier number.}:
\begin{eqnarray}
f_i(x) = p_i(x) + \epsilon_i(x) \:,
\label{eq:basic}
\end{eqnarray}
where $p_i(x)$ is the {\em a posteriori} probability distribution
of the $i$th class given input $x$, and $\epsilon_i(x)$ is the 
error associated with the $i$th output\footnote{Here, $p_i(x)$ is 
used for simplicity to denote $p(C_i|x)$.}.

\begin{figure}[htb]
\centerline{\epsfbox{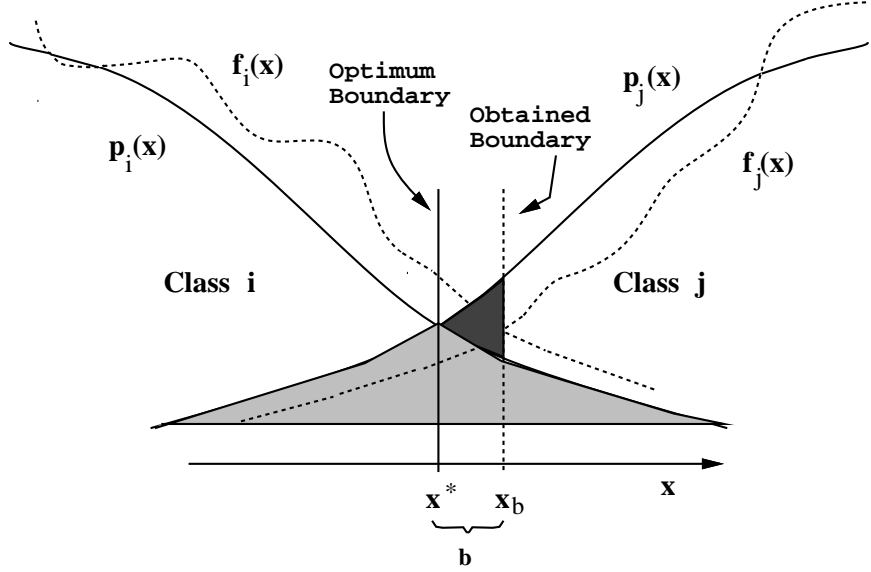}}
  \caption{Error regions associated with approximating the
    {\em a posteriori} probabilities. Lightly shaded region
    represents the Bayes error, while the darkly shaded area 
    represents the additional error due to classifier $f$.}
  \label{fig:bound3}
\end{figure}

For the Bayes optimum decision, a vector $x$ is assigned to class $i$ if 
$p_i(x) > p_k(x) \: , \; \forall k \neq i$. Therefore, the Bayes optimum 
boundary is the loci of all points $x^*$ such that 
$p_i(x^*) = p_j(x^*)$ where $ p_j(x^*) \: = \:  \max_{k \neq i} \; p_k(x) $. 
Since our classifier provides $f_i(\cdot)$ instead of
$p_i(\cdot)$, the decision boundary obtained, $x_b$, 
may vary from the optimum boundary (see Figure~\ref{fig:bound3}). 
Let $b$ denote the amount by which 
the boundary of the classifier differs from the optimum boundary 
($b = x_b - x^*$). We have:
\begin{eqnarray*}
f_i(x^* + b) = f_j(x^* + b) ,
\end{eqnarray*}
by definition of the boundary. This implies:
\begin{eqnarray}
p_i(x^* + b) + \epsilon_i(x_b) \: = \: p_j(x^* + b) + \epsilon_j(x_b) \: .
\label{eq:bd1}
\end{eqnarray}

Within a suitably chosen region about the optimum boundary, the 
{\em a posteriori} probability of the correct class
monotonically increases relative to the others as we move away from the
boundary.
This suggests a linear approximation of $p_k(x)$ around $x^*$:
\begin{eqnarray}
p_k(x^* + b) \simeq p_k(x^*) \; + \; b \: p^\prime_k(x^*) \; , \; \forall k \; ,
\label{eq:linap}
\end{eqnarray}
where $p^\prime_k(\cdot)$ denotes the derivative of $p_k(\cdot)$. With
this substitution, Equation~\ref{eq:bd1} becomes:
\begin{eqnarray}
p_i(x^*) \: + \: b \: p^\prime_i(x^*) \: + \: \epsilon_i(x_b) \; = \; 
p_j(x^*) \: + \: b \: p^\prime_j(x^*) \: + \: \epsilon_j(x_b) .
\label{eq:bd2}
\end{eqnarray}
Now, since $p_i(x^*) = p_j(x^*)$, we get:
\begin{eqnarray*}
b \: (p^\prime_j(x^*) \: - \: p^\prime_i(x^*) )
\: = \: \epsilon_i(x_b) - \epsilon_j(x_b).
\end{eqnarray*}
Finally we obtain:
\begin{eqnarray}
b \: = \: \frac{\epsilon_i(x_b) - \epsilon_j(x_b)}{s},
\label{eq:bm}
\end{eqnarray}
where:
\begin{eqnarray}
s \: = \: p^\prime_j(x^*) \: - \: p^\prime_i(x^*).
\label{eq:s}
\end{eqnarray}

Let the error $\epsilon_i(x_b)$ be broken into a bias and noise
term ($\epsilon_i(x_b) = \beta_i + \eta_i(x_b))$.
Note that the term ``bias'' and ``noise'' are only analogies, since
the error is due to the classifier as well as the data.
For the time being, the bias is assumed to be zero (i.e. $\beta_k = 0 \;
\forall k$).
The case with nonzero bias will be discussed at the end of this section.
Let $\sigma^2_{\eta_k}$ denote the variances of $\eta_k(x)$, which are taken
to be i.i.d. variables\footnote{Each output of
each network does approximate a smooth function, and therefore  the noise
for two nearby patterns on the same class (i.e. $\eta_k(x)$ and
$\eta_k(x+\Delta x)$) is correlated.
The independence assumption applies to inter-class noise
(i.e. $\eta_i(x)$ and $\eta_j(x)$), not intra-class noise.}.
Then, the variance of the zero-mean variable $b$ is given by 
(using Equation~\ref{eq:bm}):
\begin{eqnarray}
\sigma^2_b \: =\:  \frac{2 \: \sigma^2_{\eta_k}}{s^2}.
\label{eq:varb}
\end{eqnarray}

Figure~\ref{fig:bound3} shows the {\em a posteriori} probabilities obtained
by a non-ideal classifier, and the associated added error region.
The lightly shaded area provides the Bayesian error region. The darkly
shaded area is the added error region associated with selecting a decision
boundary that is offset by $b$, 
since patterns corresponding to the darkly shaded region are erroneously 
assigned to class $i$ by the classifier, although ideally they should 
be assigned to class $j$.  

The added error region, denoted by
$A(b)$, is given by:
\begin{eqnarray}
A(b) \: = \: \int_{x^*}^{x^*+b} \left( p_j(x) - p_i(x) \right)dx.
\label{eq:arint}
\end{eqnarray}
Based on this area, the expected added error, $E_{add}$, is given by:
\begin{eqnarray}
E_{add} = \int_{-\infty}^\infty A(b) f_b(b) db,
\label{eq:err}
\end{eqnarray}
where $f_b$ is the density function for $b$.
More explicitly, the expected added error is:
\begin{eqnarray*}
E_{add} = \int_{-\infty}^\infty \! \int_{x^*}^{x^*+b} \left( p_j(x) - p_i(x)
\right) \: f_b(b) \:  dx db.
\end{eqnarray*}

One can compute $A(b)$ directly by using the approximation in
Equation~\ref{eq:linap} and solving Equation~\ref{eq:arint}.
The accuracy of this
approximation depends on the proximity of the boundary to the ideal boundary.
However, since in general, the boundary density decreases rapidly
with increasing distance from the ideal boundary, 
the approximation is reasonable at least
for the most likely (i.e. small) values of $b$.
This leads to: 
\begin{eqnarray*}
A(b)  =  \int_{x^*}^{x^*+b} 
\! \! \! \! \left( (p_j(x^*)  \; + \;  (x-x^*) \: p^\prime_j(x^*)) \; - \;
(p_i(x^*)  \; + \;  (x-x^*) \: p^\prime_i(x^*) )  \right)dx.
\label{eq:intap}
\end{eqnarray*}
or:
\begin{eqnarray}
A(b)  =  \frac{1}{2} \: b^2 \: s,
\label{eq:area}
\end{eqnarray}
where $s$ is given by Equation~\ref{eq:s}.

Equation~\ref{eq:err} shows how the error can be obtained directly
from the density function of the boundary offset. 
Although obtaining the exact form of the density function for $b$ is
possible (it is straightforward for linear combiners, but
convoluted for order statistics combiners), it is not required. 
Since the area given in Equation~\ref{eq:area} is a polynomial of the
second degree, we can find its expected value using the first two moments
of the distribution of $b$.
Let us define the first and second moments of the
the boundary offset:
\begin{eqnarray*}
M_1 \: =\: \int_{-\infty}^\infty x f_b(x) dx.
\end{eqnarray*}
and:
\begin{eqnarray*}
M_2 \: =\: \int_{-\infty}^\infty x^2 f_b(x) dx.
\end{eqnarray*}

Computing the expected error for a combiner reduces to solving:
\begin{eqnarray*}
E_{add} = \int_{-\infty}^\infty \frac{1}{2} b^2 s f_b(b) db,
\end{eqnarray*}
in terms of $M_1$ and $M_2$, leading to:
\begin{eqnarray}
E_{add} = \frac{s}{2}\int_{-\infty}^\infty b^2 f_b(b) db = 
\frac{s M_2}{2}.
\label{eq:errm2}
\end{eqnarray}

The offset $b$ of a single classifier without
bias has $M_1 = 0$ and $M_2 = \sigma^2_b$, leading to: 
\begin{eqnarray}
E_{add} = \frac{s \sigma^2_b}{2}.
\label{eq:errsing}
\end{eqnarray}
Of course, Equation~\ref{eq:errsing} only provides the added error. The
total error is the sum of the added error and the Bayes error, which is 
given by:
\begin{eqnarray}
E_{tot} & = & E_{bay}  +  E_{add} .
\end{eqnarray}

Now, if the classifiers are biased, 
we need to proceed with the assumption that
$\epsilon_k(x) = \beta_k + \eta_k(x)$ where $\beta_k \neq 0$.
The boundary offset for
a single classifier becomes:
\begin{eqnarray}
b \: = \: \frac{\eta_i(x_b) - \eta_j(x_b)}{s} + \frac{\beta_i - \beta_j}{s}.
\end{eqnarray}
In this case, the variance of $b$ is left unchanged (given by 
Equation~\ref{eq:varb}), but the mean becomes
$\beta = \frac{\beta_i - \beta_j}{s}$. In other words, we have 
$M_1 = \beta$ and $\sigma^2_b = M_2 - {M_1}^2$, leading to the 
following added error:
\begin{eqnarray}
E_{add}(\beta) = \frac{s M_2}{2} \; = \; \frac{s}{2} \: (\sigma^2_b + \beta^2).
\label{eq:bising}
\end{eqnarray}

For analyzing the error regions after combining and comparing
them to the single classifier case, one needs to determine how 
the first and second moment of the boundary 
distributions are affected by combining.
The bulk of the work in the following sections focuses on obtaining
those values.

\section{Linear Combining}
\label{sec:lin}
\subsection{Linear Combining of Unbiased Classifiers}
\label{sec:unlin}
Let us now divert our attention to the effects of linearly combining
multiple classifiers. In what follows, the combiner denoted
by $ave$ performs an arithmetic average in output space. 
If $N$ classifiers are available, the $i$th output of the $ave$ combiner 
provides an approximation to $p_i(x)$ given by:
\begin{eqnarray}
f_i^{ave}(x) = \frac{1}{N} \; \sum_{m=1}^N f_i^m(x), 
\label{eq:ave}
\end{eqnarray}
or:
\begin{eqnarray*}
f_i^{ave}(x) = p_i(x) \: + \: \bar{\beta}_i \: + \: \bar{\eta}_i(x) \:,
\end{eqnarray*}
where:
\begin{eqnarray*}
\bar{\eta}_i(x) = \frac {1}{N} \;\sum_{m=1}^N \eta_i^m(x) \; ,
\end{eqnarray*}
and
\begin{eqnarray*}
\bar{\beta}_i = \frac {1}{N} \;\sum_{m=1}^N \beta^m_i \: . 
\end{eqnarray*}
If the classifiers are unbiased, $\bar{\beta}_i \: = \:  0$. Moreover,
if the errors of different classifiers are i.i.d.,
the variance of $\bar{\eta}_i$ is given by:
\begin{eqnarray}
\sigma^2_{\bar{\eta_i}} = \frac{1}{N^2} \sum_{m=1}^N \sigma^2_{\eta^m_i}
\; = \; \frac{1}{N} \sigma^2_{\eta_i}  \:.
\label{eq:varred}
\end{eqnarray}
The boundary $x^{ave}$ then has an offset $b^{ave}$, where:
\begin{eqnarray*}
f_i^{ave}(x^* + b^{ave}) = f^{ave}_j(x^* + b^{ave}) ,
\end{eqnarray*}
and: 
\begin{eqnarray}
b^{ave} \: = \:
\frac{\bar{\eta_i}(x_{b^{ave}}) - \bar{\eta_j}(x_{b^{ave}})}{s}. 
\label{eq:bave}
\end{eqnarray}
The variance of $b^{ave}$, $\sigma^2_{b^{ave}}$, can be computed in a manner
similar to $\sigma^2_{b}$, resulting in:
\begin{eqnarray*}
\sigma^2_{b^{ave}} \: = \: 
\frac{\sigma^2_{\bar{\eta_i}} + \sigma^2_{\bar{\eta_j}}}{s^2},
\end{eqnarray*}
which, using Equation~\ref{eq:varred}, leads to:
\begin{eqnarray*}
\sigma^2_{b^{ave}} \: = \: 
\frac{\sigma^2_{\eta_i} + \sigma^2_{\eta_j}} {N \: s^2},
\end{eqnarray*}
or:
\begin{eqnarray}
\sigma^2_{b^{ave}} \: = \: 
\frac{\sigma^2_b} {N}. 
\label{eq:var}
\end{eqnarray}

Qualitatively, this reduction in variance
can be readily translated into a reduction in error rates, since
a narrower boundary distribution means 
the likelihood that a boundary will be near the ideal one is increased.
In effect, using the evidence of more than one classifier 
reduces the variance of the class boundary, 
thereby providing a ``tighter'' error-prone area.
In order to establish the exact improvements in the classification
rate, we need to compute the expected added error region, and explore the
relationship between classifier boundary variance and error rates.

To that end, let us return to the added error region analysis. 
For the $ave$ classifier, the first and second moments of the boundary
offset, $b^{ave}$, are: 
$M_1^{ave} = 0$ and $M_2^{ave} = \sigma^2_{b^{ave}}$.
Using Equation~\ref{eq:var}, we obtain $M_2^{ave} = \frac{\sigma^2_b} {N}$.
The added error for the $ave$ combiner becomes:
\begin{eqnarray}
E_{add}^{ave} = \frac{s M_2^{ave}}{2} \; = \; \frac{s}{2} \: \sigma^2_{b^{ave}}
\; = \; \frac{s}{2 \: N} \: \sigma^2_b  \; = \; \frac{E_{add}}{N} .
\label{eq:errave}
\end{eqnarray}

Equation~\ref{eq:errave} quantifies the improvements due to combining
$N$ classifiers.
Under the assumptions discussed above, combining
in output space reduces added error regions by a factor
of $N$. Of course, the total error, which is the sum of Bayes error
and the added error, will be reduced by a smaller amount, since
Bayesian error will be non-zero for problems with overlapping classes.
In fact, this result, coupled with the reduction factor obtained is
Section~\ref{sec:comcor},  can be used to provide
estimates for the Bayes error rate \cite{tugh95e,tugh96c}.

\subsection{Linear Combining of Biased Classifiers}
\label{sec:bilin}
In general, $\bar{\beta}_i$ is nonzero since at least one classifier is
biased. In this case, the boundary offset $b^{ave}$ becomes:
\begin{eqnarray}
b^{ave} \: = \: 
\frac{\bar{\eta_i}(x_{b^{ave}}) - \bar{\eta_j}(x_{b^{ave}})}{s} + 
\frac{\bar{\beta_i} - \bar{\beta_j}}{s}.
\end{eqnarray}
The variance of  $\bar{\eta}_i(x)$ is identical to that of the unbiased
case, but the mean of $b^{ave}$ is given by $\bar{\beta}$ where
\begin{eqnarray}
\bar{\beta} =  \frac{\bar{\beta_i} - \bar{\beta_j}}{s}.
\label{eq:barbeta}
\end{eqnarray}
The effect of combining is less clear in this case, since the average bias 
($\bar{\beta}$) is not necessarily less than each of the individual biases. 
Let us determine the first and second moments of $b^{ave}$.
We have $M_1^{ave} = \bar{\beta}$, and 
$\sigma^2_{b^{ave}} = M_2^{ave} - (M_1^{ave})^2$, leading to:
\begin{eqnarray*}
E_{add}^{ave}(\bar{\beta}) \; = \; \frac{s M_2^{ave}}{2} 
\; = \;  \frac{s}{2} \; ( \sigma^2_{b^{ave}} + (\bar{\beta})^2 )
\end{eqnarray*}
which is:
\begin{eqnarray}
E_{add}^{ave}(\bar{\beta}) \; = \;
\frac{s}{2} \; \left( \frac{\sigma^2_b}{N} + \frac{\beta^2}{z^2} \right)
\label{eq:bired}
\end{eqnarray}
where $\bar{\beta} = \frac{\beta}{z}$, and $z \geq 1$. Now let us
limit the study to the case where $z \leq \sqrt{N}$. 
Then\footnote{If $z \geq \sqrt{N}$, then
the reduction of the variance becomes the limiting factor, and the
reductions established in the previous section hold.}:
\begin{eqnarray*}
E_{add}^{ave}(\bar{\beta}) \; \leq \;
\frac{s}{2} \; \left( \frac{\sigma^2_b + \beta^2}{z^2} \right)
\end{eqnarray*}
leading to:
\begin{eqnarray}
E_{add}^{ave}(\bar{\beta}) \; \leq \;
\frac{1}{z^2} \: E_{add}(\beta).
\label{eq:redbias}
\end{eqnarray}

Equation~\ref{eq:redbias} quantifies the error reduction in the
presence of network bias. The improvements are more modest than
those of the previous section, since both the bias and the 
variance of the noise need to be reduced. If both the variance and
the bias contribute to the error, and their contributions are
of similar magnitude, the actual reduction
is given by $min(z^2,N)$. If the bias can be kept
low (e.g. by purposefully using a larger network than required), then
once again $N$ becomes the reduction factor. 
These results highlight the basic strengths of combining, which not
only provides improved error rates, but is also a method of controlling
the bias and variance components of the error separately, thus providing
an interesting solution to the bias/variance problem~\cite{gebi92}. 

\section{Order Statistics}
\label{sec:order}
\subsection{Introduction}
Approaches to pooling classifiers can be separated
into two main categories: simple combiners, e.g.,
averaging,
and computationally expensive combiners, e.g., stacking.
The simple combining methods are
best suited for problems
where the individual classifiers perform the same task, and have
comparable success. However, such combiners are susceptible to outliers
and to unevenly performing classifiers.
In the second category,
``meta-learners,'' i.e., either sets of combining rules, or
full fledged classifiers acting on the outputs of
the individual classifiers, are constructed.
This type of
combining is more general, but suffers from all the problems associated
with the extra learning (e.g., overparameterizing, lengthy training time).

Both these methods are in fact ill-suited for problems where {\em most}
(but not all)
classifiers perform within a well-specified range. In such cases
the simplicity of averaging the classifier outputs is appealing,
but the prospect of one poor classifier corrupting the combiner
makes this a risky choice.
Although, weighted averaging of classifier outputs appears to provide some
flexibility, obtaining the optimal weights can be computationally
expensive.
Furthermore, the weights are generally assigned on a per
classifier, rather than per sample or per class basis.
If a classifier is accurate only in
certain areas of the inputs space, this scheme fails to take
advantage of the variable accuracy of the classifier in question.
Using a meta learner that would have
weights for each classifier on each pattern, would solve this problem,
but at a considerable cost.
The robust combiners presented in this
section aim at bridging the gap between simplicity and generality
by allowing the flexible selection of classifiers
without the associated cost of training meta classifiers.

\subsection{Background}
In this section we will briefly discuss some basic concepts and
properties of order statistics. Let $X$ be a random variable
with a probability density function $f_X(\cdot)$, and cumulative distribution
function $F_X(\cdot)$. Let $(X_1,X_2,\cdots,X_N)$ be a random sample drawn
from this distribution.
Now, let us arrange them in non-decreasing order, providing: 
\begin{eqnarray*}
X_{1:N} \leq  X_{2:N} \leq \; \cdots \; \leq X_{N:N}.
\end{eqnarray*}
The $i$th order statistic denoted by $X_{i:N}$, is the $i$th value
in this progression. The cumulative distribution function for the smallest
and largest order statistic can be obtained by noting that:
\begin{eqnarray*}
F_{X_{N:N}}(x) = P(X_{N:N} \leq x) = \Pi_{i=1}^N P(X_{i:N} \leq x) = [F_X(x)]^N \end{eqnarray*}
and:
\begin{eqnarray*}
F_{X_{1:N}}(x) & = &  P(X_{1:N} \leq x) = 1 - P(X_{1:N} \geq x) = 
1 - \Pi_{i=1}^N P(X_{i:N} \geq x) \\
& = & 1 - (1 - \Pi_{i=1}^N P(X_{i:N} \leq x) = 1 - [ 1 - F_X(x)]^N 
\end{eqnarray*}
The corresponding probability density functions can be obtained from
these equations. 
In general, for the $i$th order statistic, the cumulative distribution
function gives the probability that exactly $i$ of the chosen $X$'s are
less than or equal to $x$. 
The probability density function of $X_{i:N}$ is then
given by \cite{davi70}:
\begin{eqnarray}
f_{X_{i:N}}(x) = \frac{N!}{(i-1)! \: (N-i)!} \left[F_X(x) \right]^{i-1} 
\left[1 - F_X(x)\right]^{N-i} f_X(x) \; .    
\label{eq:osdensity}
\end{eqnarray}
This general form however, cannot always be computed in closed form. 
Therefore, obtaining
the expected value of a function of $x$ using Equation~\ref{eq:osdensity}
is not always possible. However, the first two moments of the density
function are widely available for a variety of distributions \cite{arba92}.
These moments can be used to compute the expected values of certain specific
functions, e.g. polynomials of order less than two.

\subsection{Combining Unbiased Classifiers through OS}
\label{sec:osno}
Now, let us turn our attention to order statistic combiners. 
For a given input $x$, let the network outputs of each of the $N$ 
classifiers for each class $i$ be ordered in the following manner:
\begin{eqnarray*}
f^{1:N}_i(x) \leq  f^{2:N}_i(x) \leq
\; \cdots \; \leq f^{N:N}_i(x).
\end{eqnarray*}
Then, the $max$, $med$ and $min$ combiners are defined 
as follows \cite{davi70}:
\begin{eqnarray}
\label{eq:max}
f_i^{max}(x)\; & = & \; f^{N:N}_i(x), \\
\label{eq:med}
f_i^{med}(x)\; & = & \; \left\{ \begin{array}{ll}
     \frac {f^{\frac{N}{2}:N}_i(x) \; + \:
f^{\frac{N}{2}+1:N}_i(x)}{2} & \mbox{if $N$ is even} \\
f^{\frac{N+1}{2}:N}_i(x)           & \mbox{if $N$ is odd},
                       \end{array} \right. \\
\label{eq:min}
f_i^{min}(x)\; & = & \; f^{1:N}_i(x). 
\end{eqnarray}
These three combiners are chosen because they represent important
qualitative interpretations of the output space. Selecting the maximum
combiner is equivalent to selecting the class with the highest
posterior. Indeed, since the network outputs approximate the
class {\em a posteriori} distributions, selecting the maximum 
reduces to selecting the classifier with the highest confidence
in its decision. The drawback of this method however is that it can
be compromised by a single classifier that repeatedly provides high
values. The selection of the minimum combiner follows a similar logic, but 
focuses on classes that are unlikely to be correct, rather than on the
correct class. Thus, this combiner eliminates less likely classes by basing 
the decision on the lowest value for a given class. This combiner suffers
from the same ills as the {\em max} combiner, although it is 
less dependent on a single error, since it performs a min-max operation,
rather than a max-max\footnote{Recall that the pattern is ultimately
assigned to the class with the highest combined output.}.  
The median classifier on the other hand considers the most ``typical'' 
representation of each class. For highly
noisy data, this combiner is more desirable than either the {\em min}
or {\em max} combiners since the decision is not compromised as much
by a single large error.

The analysis of the properties of these combiners does not depend
on the order statistic chosen. 
Therefore we will denote all three by $f_i^{os}(x)$
and derive the error regions. 
The network output provided by
$f_i^{os}(x)$ is given by:
\begin{eqnarray}
f^{os}_i(x) = p_i(x) + \epsilon^{os}_i(x) \:,
\label{eq:os}
\end{eqnarray}

Let us first investigate the zero-bias case ($\beta_k = 0 \: \forall k$). 
We get $\epsilon^{os}_k(x) = \eta^{os}_k(x) \:
\forall k$, since the variations in the $k$th output of the classifiers
are solely due to noise. 
Proceeding as before, the boundary $b^{os}$ is shown to be:
\begin{eqnarray}
b^{os} \: = \: \frac{\eta^{os}_i(x_b) - \eta^{os}_j(x_b)}{s}.
\label{eq:bos}
\end{eqnarray}
Since $\eta_k$'s are i.i.d, and $\eta_k^{os}$ is the same order statistic 
for each class, the moments will be identical for each class. Moreover,
taking the
order statistic will shift the mean of both $\eta^{os}_i$ and $\eta^{os}_j$
by the same amount, leaving the mean of the difference unaffected. 
Therefore, $b^{os}$ will have zero mean, and variance:
\begin{eqnarray}
\sigma^2_{b^{os}} \: =\: \frac{2 \: \sigma^2_{\eta^{os}_k}}{s^2}
\: =\: \frac{2 \: \alpha \sigma^2_{\eta_k}}{s^2} \: =\: \alpha \sigma^2_b,
\label{eq:osvar}
\end{eqnarray}
where $\alpha$ is a reduction factor that depends on the order 
statistic and on the distribution of $b$. For most distributions, 
$\alpha$ can be found in tabulated form \cite{arba92}.
For example, Table~\ref{tab:alpha} 
provides $\alpha$ values for all three {\em os} combiners, up to $15$ 
classifiers, for a Gaussian distribution \cite{arba92,sagr56}.

Returning to the error calculation, we have: $M_1^{os} = 0$, and 
$M_2^{os} = \sigma^2_{b^{os}}$, 
providing: 
\begin{eqnarray}
E_{add}^{os} = \frac{s M_2^{os}}{2} = 
\frac{s \sigma^2_{b^{os}}}{2} = 
\frac{s \alpha \sigma^2_b}{2} = \alpha \; E_{add}.
\label{eq:red}
\end{eqnarray}

\begin{table} [htb] \centering
  \caption{Reduction factors $\alpha$ for the {\em min, max} and {\em med} 
           combiners.}
  \begin{tabular}{|c|c|c|} \hline
N  &\multicolumn{2}{c|}{OS Combiners} \\ \cline{2-3}
   & minimum/maximum & {  median }  \\ \hline
1  &  1.00  & 1.00 \\
2  &  .682  & .682 \\
3  &  .560  & .449 \\
4  &  .492  & .361 \\
5  &  .448  & .287 \\
6  &  .416  & .246 \\
7  &  .392  & .210 \\
8  &  .373  & .187 \\
9  &  .357  & .166 \\
10 &  .344  & .151 \\
11 &  .333  & .137 \\
12 &  .327  & .127 \\
13 &  .315  & .117 \\
14 &  .308  & .109 \\
15 &  .301  & .102 \\ \hline
  \end{tabular}
\label{tab:alpha}
\end{table}

Equation~\ref{eq:red} shows that the reduction in the error region is 
directly related to the
reduction in the variance of the boundary offset $b$. Since the means and
variances of order statistics for a variety of distributions are widely 
available in tabular form, the reductions can be readily quantified.

\subsection{Combining Biased Classifiers through OS}
\label{sec:osbi}
In this section, we analyze the error regions in the presence of bias.
Let us study $b^{os}$ in detail when multiple classifiers are combined
using order statistics. First 
note that the bias and noise cannot be separated, since in general
$(a + b)^{os} \neq a^{os} + b^{os}$. We will therefore need to specify 
the mean and variance of the result of each operation\footnote{Since the
exact distribution parameters of $b^{os}$ are not known, 
we use the sample mean and the sample variance.}.
Equation~\ref{eq:bos} becomes:
\begin{eqnarray}
b^{os} \: = \: \frac{(\beta_i + \eta_i(x_b))^{os} - 
(\beta_j + \eta_j(x_b))^{os}}{s}.
\end{eqnarray}
Now, $\beta_k$ has mean $\bar{\beta_k}$, given 
by $\frac {1}{N} \;\sum_{m=1}^N \beta^m_k \:$ , 
where $m$ denotes the different classifiers.
Since the noise is zero-mean, $\beta_k + \eta_k(x_b)$ has first 
moment $\bar{\beta_k}$ and variance
$\sigma^2_{\eta_k} + \sigma^2_{\beta_k}$, where $\sigma^2_{\beta_k} =
\frac {1}{N-1} \; \sum_{m=1}^N (\beta^m_k - \bar{\beta_k})^2 \: $. 

Taking a specific order statistic of this
expression will modify both moments. The first moment is given
by $\bar{\beta_k} + \mu^{os}$, where $\mu^{os}$ is a shift which depends 
on the order statistic chosen, but not on the class. 
The first moment of $b^{os}$ then, is given by:
\begin{eqnarray}
\frac{(\bar{\beta_i} + \mu^{os})  - (\bar{\beta_j} + \mu^{os})}{s} \: = 
\frac{\bar{\beta_i} - \bar{\beta_j}}{s} \: = \: \bar{\beta}.
\end{eqnarray}
Note that the bias term represents an ``average bias'' since the
contributions due to the order statistic are removed. Therefore,
reductions in bias cannot be obtained 
from a table similar to Table~\ref{tab:alpha}.  

Now, let us turn our attention to the variance. 
Since $\beta_k + \eta_k(x_b)$ has variance 
$\sigma^2_{\eta_k} + \sigma^2_{\beta_k}$, it
follows that $(\beta_k + \eta_k(x_b))^{os}$ has variance 
$\sigma^2_{\eta_k^{os}} = \alpha (\sigma^2_{\eta_k} + \sigma^2_{\beta_k})$,
where $\alpha$ is the factor discussed in Section~\ref{sec:osno}. 
Since $b^{os}$ is a linear combination of error terms, its variance is
given by:
\begin{eqnarray}
\sigma^2_{b^{os}} & =& \frac{\sigma^2_{\eta^{os}_i} + \sigma^2_{\eta^{os}_j}}
{s^2}
\: =\: \frac{2 \: \alpha \sigma^2_{\eta_i}}{s^2} +  
 \frac{\alpha (\sigma^2_{\beta_i} +  \sigma^2_{\beta_j})}{s^2}  \\ 
& =& \alpha (\sigma^2_b + \sigma^2_{\beta}),
\end{eqnarray}
where $\sigma^2_{\beta} = \frac{\sigma^2_{\beta_i} + \sigma^2_{\beta_j}}{s^2}$
is the variance introduced by the biases of different classifiers.
The result of bias then manifests itself both in the mean and the
variance of the boundary offset $b^{os}$.

We have now obtained the first and second moments of $b^{os}$,
and can compute the added error region. Namely, we have 
$M_1^{os}\: = \: \bar{\beta}$ and 
$\sigma^2_{b^{os}} \: = \: M_2^{os}- (M_1^{os})^2$ leading to:
\begin{eqnarray}
E^{os}_{add}(\beta) & = & 
\frac{s}{2} \: M_2^{os} \: = \: \frac{s}{2} \: 
(\sigma^2_{b^{os}} + \bar{\beta}^2) \\
& = & \frac{s}{2} \: (\alpha (\sigma^2_b + \sigma^2_{\beta})+ \bar{\beta}^2).
\label{eq:addbi}
\end{eqnarray}
The reduction in the error is more difficult to assess in this case. 
By writing the error as:
\begin{eqnarray*}
E_{add}^{os}(\beta) \: = \:  
\alpha (\frac{s}{2} \: (\sigma^2_b + {\beta}^2)) \: + \:  
\frac{s}{2} \: (\alpha \sigma^2_{\beta} + \bar{\beta}^2 - \alpha {\beta}^2), 
\end{eqnarray*}
we get:
\begin{eqnarray}
E_{add}^{os}(\beta) \: = \: \alpha \: E_{add}(\beta) + 
\frac{s}{2} \: (\alpha \sigma^2_{\beta} + \bar{\beta}^2 - \alpha {\beta}^2) .
\label{eq:addbi2}
\end{eqnarray}
Analyzing the error reduction in the general case requires knowledge
about the bias introduced by each classifier. However, it is possible
to analyze the extreme cases. If each classifier has the same bias
for example, $\sigma^2_{\beta}$ is reduced to zero and $\bar{\beta} = \beta$.
In this case the error reduction can be expressed as:
\begin{eqnarray*}
E_{add}^{os}(\beta) \: = \: 
\frac{s}{2} \: (\alpha \sigma^2_b + {\beta}^2), 
\end{eqnarray*}
where only the error contribution due to the variance of $b$ is reduced.
In this case it is important to reduce classifier bias before
combining (e.g. by using an overparametrized model).
If on the other hand, the biases produce a zero mean variable, i.e. they
cancel each other out, we obtain $\bar{\beta} = 0$. In this case, the added 
error becomes:
\begin{eqnarray*}
E_{add}^{os}(\beta) \: = \: \alpha \: E_{add}(\beta) +
\frac{s \: \alpha}{2} \: (\sigma^2_{\beta} - {\beta}^2)
\end{eqnarray*}
and the error reduction will be significant as long as
$\sigma^2_{\beta} \leq {\beta}^2$.

\section{Correlated Classifier Combining}
\label{sec:indep}
\subsection{Introduction}
The discussion so far focused on finding the
types of combiners that improve performance.
Yet, it is important to
note that if the classifiers to be combined repeatedly provide the
same (either erroneous or correct) classification decisions, there
is little to be gained from combining, regardless of the chosen scheme.
Therefore, the selection and training
of the classifiers that will be combined is as critical an issue as
the selection of the combining method. Indeed, classifier/data
selection is directly tied to the amount of correlation among the
various classifiers, which in turn affects the amount
of error reduction that can be achieved.

The tie between error correlation and classifier performance
was directly or indirectly observed by many researchers.
For regression problems, Perrone and Cooper show that their
combining results are weakened if the
networks are not independent~\cite{peco93}. Ali and Pazzani discuss
the relationship between error correlations and error reductions in
the context of decision trees~\cite{alpa95a}.
Meir discusses the effect of independence on combiner
performance~\cite{meir95}, and
Jacobs reports that $N' \leq N$ independent classifiers are worth as much
as $N$ dependent classifiers~\cite{jaco95}.
The influence of the amount of training on ensemble
performance is studied in~\cite{sokr96}.
For classification problems, the effect of the correlation among
the classifier errors on combiner performance was quantified
by the authors~\cite{tugh96b}.

\subsection{Combining Unbiased Correlated Classifiers}
\label{sec:comcor}
In this section we derive the explicit relationship between
the correlation among classifier errors and the error reduction
due to combining. Let us focus on the linear combination of
unbiased classifiers.
Without the independence assumption, 
the variance of $\bar{\eta}_i$ is given by:
\begin{eqnarray*}
\sigma^2_{\bar{\eta_i}} & = & \frac{1}{N^2} \sum_{l=1}^N \sum_{m=1}^N 
cov(\eta^m_i(x),\eta^l_i(x)) \\
& = & \frac{1}{N^2} \sum_{m=1}^N \sigma^2_{\eta^m_i(x)} 
+ \frac{1}{N^2} \sum_{m=1}^N \sum_{l \neq m} cov(\eta^m_i(x),\eta^l_i(x)) 
\end{eqnarray*}
where $cov(\cdot,\cdot)$ represents the covariance. 
Expressing the covariances in term of the correlations ($cov(x,y) \: = \: 
corr(x,y) \: \sigma_x \: \sigma_y$), leads to:
\begin{eqnarray}
\sigma^2_{\bar{\eta_i}}  =  \frac{1}{N^2} \sum_{m=1}^N \sigma^2_{\eta^m_i(x)} 
+ \frac{1}{N^2} \sum_{m=1}^N \sum_{l \neq m} corr(\eta^m_i(x),\eta^l_i(x))
\sigma_{\eta^m_i(x)} \sigma_{\eta^l_i(x)}.
\label{eq:siggen}
\end{eqnarray}

In situations where the variance of a given output is comparable 
across the different classifiers, 
Equation~\ref{eq:siggen} is significantly simplified by using the common
variance $\sigma_{\eta_i}$, thus becoming:
\begin{eqnarray*}
\sigma^2_{\bar{\eta_i}} & = & \frac{1}{N} \sigma^2_{\eta_i(x)}
+ \frac{1}{N^2} \sum_{m=1}^{N} \sum_{l \neq m} 
corr(\eta^m_i(x),\eta^l_i(x)) \sigma^2_{\eta_i(x)} \; . 
\end{eqnarray*}
Let $\delta_i$ be the correlation factor among all classifiers
for the $i$th output:
\begin{eqnarray*}
\delta_i = \frac {1}{N \: (N-1)} \sum_{m = 1}^N
\sum_{m \neq l} corr(\eta^m_i(x),\eta^l_i(x)).
\end{eqnarray*}
The variance of $\bar{\eta_i}$ becomes:
\begin{eqnarray*}
\sigma^2_{\bar{\eta_i}} \: = \: \frac{1}{N} \sigma^2_{\eta_i(x)}
\: + \: \frac{N-1}{N} \: \delta_i \: \sigma^2_{\eta_i(x)} \; .
\end{eqnarray*}
Now, let us return to the boundary $x^{ave}$, and its offset $b^{ave}$, where:
\begin{eqnarray*}
f_i^{ave}(x^* + b^{ave}) = f^{ave}_j(x^* + b^{ave}) .
\end{eqnarray*}
In Section~\ref{sec:unlin}, the variance of $b^{ave}$ was shown to
be:
\begin{eqnarray*}
\sigma^2_{b^{ave}} \: = \: 
\frac{\sigma^2_{\bar{\eta_i}} + \sigma^2_{\bar{\eta_j}}}{s^2}.
\end{eqnarray*}
Therefore:
\begin{eqnarray*}
\sigma^2_{b^{ave}} \: = \: \frac{1}{s^2}
\left( \frac{1}{N} \sigma^2_{\eta_i(x)} (1 + (N-1) \delta_i) \; +
\; \frac{1}{N}  \sigma^2_{\eta_j(x)} (1 + (N-1) \delta_j) \right) 
\end{eqnarray*}
which leads to:
\begin{eqnarray*}
\sigma^2_{b^{ave}} \: = \: \frac{1}{s^2 N}
\left( \sigma^2_{\eta_i(x)} \: + \:\sigma^2_{\eta_j(x)} \: + \:
(N-1) \: (\delta_i \: \sigma^2_{\eta_i(x)}  \: + \:
          \delta_j \: \sigma^2_{\eta_j(x)} )  \right),
\end{eqnarray*}
or:
\begin{eqnarray}
\sigma^2_{b^{ave}} \: = \: 
\frac{\sigma^2_{\eta_i(x)} \: + \: \sigma^2_{\eta_j(x)}}{N s^2} \; + \;
\frac{N-1}{N s^2}(\delta_i \sigma^2_{\eta_i(x)} + \delta_j \sigma^2_{\eta_j(x)}). 
\end{eqnarray}
Recalling that the noise between classes are i.i.d. leads to\footnote{The 
errors between classifiers are correlated, not the errors between classes.}:
\begin{eqnarray*}
\sigma^2_{b^{ave}} &  = & \frac{1}{N} \sigma^2_b 
\; + \: (\frac{N-1}{N}) \frac{2 \sigma^2_{\eta_j(x)}}{s^2} 
\frac{\delta_i + \delta_j}{2}\\
& = & \frac{\sigma^2_b}{N} \left(1 \: + \: (N-1) 
\frac{\delta_i + \delta_j}{2}\right).
\label{eq:cordel}
\end{eqnarray*}
This expression only considers the error that occur between classes
$i$ and $j$. 
In order to extend this expression to include all the boundaries, we
introduce an overall correlation term $\delta$.
Then, the added error is computed in terms of $\delta$.
The correlation among classifiers is calculated using the following
expression:
\begin{eqnarray}
\delta = \sum_{i=1}^{L} P_i \: \delta_i
\label{eq:cor}
\end{eqnarray}
where $P_i$ is the prior probability of class $i$. The correlation
contribution of each class to the overall correlation, 
is proportional to the prior probability of that class.

\begin{figure}[htb]
\epsfxsize = 4.6in \epsfbox{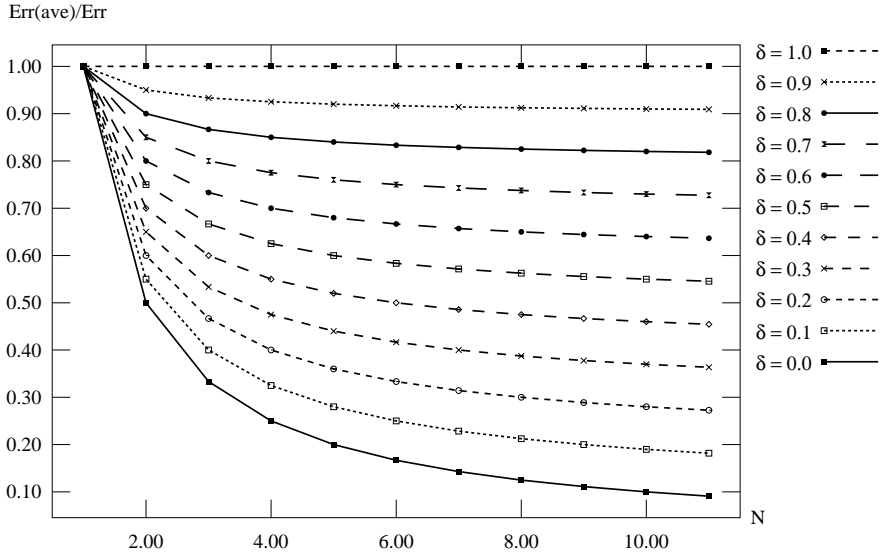}
  \caption{Error reduction ($\frac{E_{add}^{ave}} {E_{add}}$)
           for different classifier error correlations.}
  \label{fig:corr}
\end{figure}

Let us now return to the error region analysis. With this formulation
the first and second moments of $b^{ave}$ yield:
$M_1^{ave} = 0$, and $M_2^{ave} = \sigma^2_{b^{ave}}$. 
The derivation is identical to that of 
Section~\ref{sec:unlin} and the only change is in the relation between 
$\sigma^2_b$ and $\sigma^2_{b^{ave}}$. We then get:
\begin{eqnarray}
E_{add}^{ave} & = & \frac{s M_2^{ave}}{2} \; = \; 
\frac{s}{2} \: \sigma^2_{b^{ave}} \nonumber \\ \nonumber 
& = & \frac{s}{2} \: \sigma^2_b \left(\frac{1 + \delta (N - 1)}{N}\right) \\  
& = & E_{add}  \left(\frac{1 + \delta (N - 1)}{N}\right).  
\label{eq:errcor}
\end{eqnarray}

The effect of the correlation between the errors of each classifier is
readily apparent from Equation~\ref{eq:errcor}. If the errors are
independent, then the second part of the reduction term vanishes and 
the combined error is reduced by $N$. If on the other hand, the error
of each classifier has correlation $1$, then the error of the combiner is
equal to the initial errors and there is no improvement due to combining.
Figure~\ref{fig:corr} shows how the variance reduction is affected
by $N$ and $\delta$ (using Equation~\ref{eq:errcor}).
 
In general, the correlation values lie between these two extremes,
and some reduction is achieved. It is important to understand the
interaction between $N$ and $\delta$ in order to maximize the
reduction. As more and more classifiers are used (increasing $N$),
it becomes increasingly difficult to find uncorrelated classifiers.
Figure~\ref{fig:corr} can be used to determine the number of classifiers
needed for attaining satisfactory combining performance.

\subsection{Combining Biased Correlated Classifiers}
\label{sec:comcorbi}
Let us now return to the analysis of biased classifiers.
As discussed in Section~\ref{sec:lin}, the boundary offset of a single
classifier is given by:
\begin{eqnarray}
b \: = \: \frac{\eta_i(x_b) - \eta_j(x_b)}{s} + \beta \; ,
\end{eqnarray}
where $\beta = \frac{\beta_i - \beta_j}{s}$, leading to the following
added error term:
\begin{eqnarray}
E_{add}(\beta) = \frac{s}{2} \: (\sigma^2_b + \beta^2).
\label{eq:bising2}
\end{eqnarray}

Let us now focus on the effects of the $ave$ combiner on the
boundary. The combiner output is now given by:
\begin{eqnarray*}
f_i^{ave}(x) = p_i(x) \: + \: \bar{\beta}_i \: + \: \bar{\eta}_i(x) \:,
\end{eqnarray*}
where $\bar{\eta}_i(x)$ and $\bar{\beta}_i$ are given in Section~\ref{sec:lin}.
The boundary offset, $b^{ave}$ is:
\begin{eqnarray}
b^{ave} \: = \: \frac{\bar{\eta}_i(x_b) - \bar{\eta_j}(x_b)}{s} + \bar{\beta}  ,
\end{eqnarray}
where $\bar{\beta}$ is given by Equation~\ref{eq:barbeta}.
The variance of $b^{ave}$ is not affected by the biases, and the derivation
of Section~\ref{sec:comcor} applies to this case as well. 

The first and second moments of $b^{ave}$, the boundary offset obtained
using the $ave$ combiner, for biased, correlated classifiers, are given by:
$M_1^{ave} = \bar{\beta}$  and
$M_2^{ave} = \sigma^2_{b^{ave}} - (\bar{\beta})^2$.
The corresponding added error region is:
\begin{eqnarray*}
E_{add}^{ave}(\bar{\beta}) \; = \; \frac{s M_2^{ave}}{2}
\; = \;  \frac{s}{2} \; ( \sigma^2_{b^{ave}} + (\bar{\beta})^2 ) \: .
\end{eqnarray*}
Using the overall correlation term obtained in the previous section,
we can represent this expression in terms of the boundary parameters of the 
single classifier, and the bias reduction factor $z$ introduced in
Section~\ref{sec:bilin}:
\begin{eqnarray}
E_{add}^{ave}(\bar{\beta})  \; = \;  
\frac{s}{2} \; \left( 
\sigma^2_{b} \left(\frac{1 + \delta (N - 1)}{N}\right)
+ \frac{\beta^2}{z^2} \right) \: .
\label{eq:adderrco}
\end{eqnarray}
In order to obtain the error reduction rates, let us introduce $\tau$, 
the factor that will determine the final reduction:
\begin{eqnarray}
\tau^2 = min\left(z^2,\left(\frac{1 + \delta (N - 1)}{N}\right)\right).
\label{eq:tau}
\end{eqnarray}
Now, Equation~\ref{eq:adderrco} leads to:
\begin{eqnarray}
E_{add}^{ave}(\bar{\beta})  \; \leq \;
\frac{1}{\tau} \: E_{add}(\beta) \: .  
\label{eq:errcobi}
\end{eqnarray}

Equation~\ref{eq:errcobi} shows the error reduction for correlated, biased
classifiers. As long as the biases of individual classifiers are 
reduced by a larger amount than the correlated variances, the
reduction will be similar to those in Section~\ref{sec:comcor}. However,
if the biases are not reduced, the improvement gains will not be
as significant. These results are conceptually identical to those
obtained in Section~\ref{sec:lin}, but vary in how the bias reduction
$z$ relates to $N$. In effect, the requirements on reducing $z$ are lower
than they were previously, since in the presence of bias, the 
error reduction is less than $\frac{1}{N}$. 
The practical implication of this observation is that,
even in the presence of bias, the correlation dependent variance reduction
term (given in Equation~\ref{eq:errcor}) will often be the
limiting factor, and dictate the error reductions.

\subsection{Discussion}
In this section we established the importance of the correlation 
among the errors of individual classifiers in
a combiner system. One can exploit this relationship explicitly
by reducing the correlation among classifiers that will be 
combined. Several methods have been proposed for this purpose and
many researchers are actively exploring this area~\cite{shar96a}.

Cross-validation, a statistical method aimed at estimating the
``true' error~\cite{frie94,ston74,weku91}, can also be used to
control the amount of correlation among classifiers. By only training
individual classifiers on overlapping subsets of the data, the correlation
can be reduced.
The various boosting algorithms exploit the relationship between
corrlation and error rate by training subsequent classifiers on
training patterns that have been ``selected'' by earlier
classifiers~\cite{drsc93,drco96,frsc95,scfr97}
thus reducing the correlation among them.  
Krogh and Vedelsky discuss how cross-validation can be
used to improve ensemble performance~\cite{krve95}.
Bootstrapping, or generating different training sets for
each classifier by resampling the original 
set~\cite{efro82,efro83,jadu87,weku91}, provides another
method for correlation reduction~\cite{pamu96}. 
Breiman also addresses this issue, and discusses methods aimed at
reducing the correlation among estimators~\cite{brei93,brei94}.
Twomey and Smith discuss combining and resampling in the
context of a 1-$d$ regression problem~\cite{twsm95}.
The use of principal component regression to handle multi-collinearity
while combining outputs of multiple regressors, was suggested in
\cite{mepa97}. 
Another approach to reducing the
correlation of classifiers can be found in input decimation,
or in purposefully withholding some parts of each pattern
from a given classifier~\cite{tugh96b}. 
Modifying the training of individual classifiers in order
to obtain less correlated classifiers was also explored~\cite{rose96},
and the selection
of individual classifier through a genetic algorithm is
suggested in~\cite{opsh96}.

In theory, reducing the correlation among classifiers that are
combined increases the ensemble classification rates.
In practice however, since each classifier uses a subset of the
training data, individual classifier performance can deteriorate, thus
offsetting any potential gains at the ensemble level~\cite{tugh96b}.
It is therefore crucial to reduce the correlations without
increasing the individual classifiers' error rates.

\section{Experimental Combining Results}
\label{sec:resu} 
In order to provide in depth analysis and to 
demonstrate the result on public domain data sets, 
we have divided this section into two parts. First we will provide detailed
experimental results on one difficult data set, outlining all
the relevant design steps/parameters. Then we will summarize results
on several public domain data sets taken from the UCI depository/Proben1
benchmarks~\cite{prec94}. 

\subsection{Oceanic Data Set}
\label{sec:ocean}
The experimental data set used in this section is derived from
underwater SONAR signals. 
From the original SONAR signals of four different underwater objects,
two feature sets are extracted \cite{ghde92,ghtu96}.
The first one (FS1), a 25-dimensional set, consists of Gabor wavelet
coefficients, temporal descriptors and spectral measurements.  The second
feature set (FS2), a 24-dimensional set, consists of reflection coefficients
based on both short and long time windows, and temporal descriptors.
Each set consists of 496 training and 823 test patterns.
The data is available at URL {\em http://www.lans.ece.utexas.edu}.

\subsubsection{Combining Results}
In this section we present detailed results obtained from the Oceanic
data described above.
Two types of feed forward networks,
namely a multi-layered perceptron (MLP) with a single hidden layer with $50$
units and a radial basis function (RBF) network with $50$ kernels, are used to
classify the patterns.
Table~\ref{tab:ind}
provides the test set results for individual classifier/feature set pairs.
The reported error percentages are averaged over $20$ runs.
Tables~\ref{tab:fs1} and \ref{tab:fs2} show the combining
results for each feature set. 
Combining consists of utilizing the
outputs of multiple MLPs, RBFs or an MLP/RBF mix, and performing the
operations described in Equations~\ref{eq:ave},
\ref{eq:max}, \ref{eq:med} and  \ref{eq:min}.
When combining an odd number of classifiers, the classifier with the
better performance is selected once more than the less successful one.
For example, when combining the MLP and RBF results on FS1 for 
$N=5$, three RBF networks and two MLPs are used.
Table~\ref{tab:both} shows the improvements
that are obtained if more than one feature set is available\footnote{All 
the combining results provide improvements that are
statistically significant over the individual classifiers, 
or more precisely, the hypothesis
that the two means are equal ({\em t}--test) is rejected for $\alpha = .05$.}.

\begin{table} [htb] \centering
\caption{Individual Classifier Performance on Test Set.}
\begin{tabular}{|l||c|c|} \hline
Classifier/       & Error Rate & st. dev.\\ 
{Feature Set}    & &  \\ \hline \hline
 FS1/MLP    &  7.47  & 0.44 \\
 FS1/RBF    &  6.79  & 0.41 \\
 FS2/MLP    &  9.95  & 0.74 \\
 FS2/RBF    &  10.94  & 0.93 \\ \hline
\end{tabular}
\label{tab:ind}
\end{table}

The performance of the $ave$ combiner is better than that of the $os$
combiners, especially for the second feature set (FS2).
While combining information from two different feature sets, the linear
combiner performed best with the RBF classifiers, while the {\em max}
combiner performed best with the MLP classifiers. Furthermore, using
different types of classifiers does not change the performance of the
linear combiner when qualitatively different feature sets are used.
However, for the $os$ combiners, the results do improve when
both different classifier types and different feature sets are used.

\begin{table} [htb] \centering
\caption{Combining Results for FS1.}
\begin{tabular}{|l|c||c|c||c|c||c|c||c|c||c|c||} \hline
\multicolumn{2}{|c||} {Classifier(s)}& \multicolumn{2}{c||}{Ave}& \multicolumn{2}{c||}{Med} & \multicolumn{2}{c||}{Max} & \multicolumn{2}{c||}{Min}\\ \hline
 {} &N& Error& $\sigma$ & Error & $\sigma$ & Error & $\sigma$ & Error & $\sigma$ \\ \hline \hline
    &3& 7.19 &0.29& 7.25 &0.21& 7.38 &0.37& 7.19 &0.37\\
MLP &5& 7.13 &0.27& 7.30 &0.29& 7.32 &0.41& 7.20 &0.37\\
    &7& 7.11 &0.23& 7.27 &0.29& 7.27 &0.37& 7.35 &0.30\\ \hline
    &3& 6.15 &0.30& 6.42 &0.29& 6.22 &0.34& 6.30 &0.40\\
RBF &5& 6.05 &0.20& 6.23 &0.18& 6.12 &0.34& 6.06 &0.39\\
    &7& 5.97 &0.22& 6.25 &0.20& 6.03 &0.35& 5.92 &0.31\\ \hline
    &3& 6.11 &0.34& 6.02 &0.33& 6.48 &0.43& 6.89 &0.29\\
BOTH&5& 6.11 &0.31& 5.76 &0.29& 6.59 &0.40& 6.89 &0.24\\
    &7& 6.08 &0.32& 5.67 &0.27& 6.68 &0.41& 6.90 &0.26\\ \hline
\end{tabular}
\label{tab:fs1}
\end{table}

\begin{table} [htb] \centering
\caption{Combining Results for FS2.}
\begin{tabular}{|l|c||c|c||c|c||c|c||c|c||c|c||} \hline
\multicolumn{2}{|c||} {Classifier(s)}& \multicolumn{2}{c||}{Ave}& \multicolumn{2}{c
||}{Med} & \multicolumn{2}{c||}{Max} & \multicolumn{2}{c||}{Min} \\ \hline
{}& N & Error& $\sigma$ & Error & $\sigma$ & Error & $\sigma$ & Error & $\sigma$ \\ \hline \hline
    &3 & 9.32 &0.35& 9.47 &0.47& 9.64 &0.47& 9.39 &0.34\\
MLP &5 & 9.20 &0.30& 9.22 &0.30& 9.73 &0.44& 9.27 &0.30\\
    &7 & 9.07 &0.36& 9.11 &0.29& 9.80 &0.48& 9.25 &0.36\\ \hline
    &3 &10.55 &0.45&10.49 &0.42&10.59 &0.57&10.74 &0.34\\
RBF &5 &10.43 &0.30&10.51 &0.34&10.55 &0.40&10.65 &0.37\\
    &7 &10.44 &0.32&10.46 &0.31&10.58 &0.43&10.66 &0.39\\ \hline
    &3 & 8.46 &0.57& 9.20 &0.49& 8.65 &0.47& 9.56 &0.53\\
BOTH&5 & 8.17 &0.41& 8.97 &0.54& 8.71 &0.36& 9.50 &0.45\\
    &7 & 8.14 &0.28& 8.85 &0.45& 8.79 &0.40& 9.40 &0.39\\ \hline
\end{tabular}
\label{tab:fs2}
\end{table}

\begin{table} [htb] \centering
\caption{Combining Results when Both Feature Sets are Used.}
\begin{tabular}{|l|c||c|c||c|c||c|c||c|c||c|c||} \hline
\multicolumn{2}{|c||} {Classifier(s)}& \multicolumn{2}{c||}{Ave}& \multicolumn{2}{c||}{Med} & \multicolumn{2}{c||}{Max} & \multicolumn{2}{c||}{Min} \\ \hline
{}& N  & Error& $\sigma$ & Error & $\sigma$ & Error & $\sigma$ & Error & $\sigma$ \\ \hline \hline
    &3& 5.21 &0.33& 6.25 &0.36& 4.37 &0.41& 4.72 & 0.28 \\
MLP &5& 4.63 &0.35& 5.64 &0.32& 4.22 &0.41& 4.58 & 0.17 \\
    &7& 4.20 &0.40& 5.29 &0.28& 4.13 &0.34& 4.51 & 0.20 \\ \hline
    &3& 3.70 &0.33& 5.78 &0.32& 4.76 &0.37& 3.93 &0.50 \\
RBF &5& 3.40 &0.21& 5.38 &0.38& 4.73 &0.35& 3.83 &0.43 \\
    &7& 3.42 &0.21& 5.15 &0.31& 4.70 &0.36& 3.76 &0.33 \\ \hline
    &3& 3.94 &0.24& 4.52 &0.29& 4.34 &0.42& 4.51 &0.30 \\
BOTH&5& 3.42 &0.23& 4.35 &0.32& 4.13 &0.49& 4.48 &0.29 \\
    &7& 3.40 &0.26& 4.05 &0.29& 4.10 &0.36& 4.39 &0.24 \\ \hline
\end{tabular}
\label{tab:both}
\end{table}

\subsubsection{Correlation Factors}
Let us now estimate
the correlation factors among the different classifiers in order to 
determine the compatibility of the various classifier/feature set pairs.
The data presented in Section~\ref{sec:ocean} will be used in this section.
Table~\ref{tab:corr} shows the estimated average error correlations between:
\begin{itemize}
\item{}
different runs of a single classifier on a single feature set (first four rows);
\item{} different classifiers trained with a single feature set 
(fifth and sixth rows); 
\item{} single classifier trained on two different feature sets
(seventh and eighth rows).
\end{itemize}

\begin{table} [htb] \centering
\caption{Experimental Correlation Factors Between Classifier Errors.}
\begin{tabular}{|l||c|} \hline
        Feature Set/Classifier Pairs& Estimated Correlation  \\ \hline \hline
Two runs of FS1/MLP    & 0.89 \\
Two runs of FS1/RBF    & 0.79 \\
Two runs of FS2/MLP    & 0.79 \\
Two runs of FS2/RBF    & 0.77 \\ \hline
FS1/MLP and FS1/RBF    & 0.38 \\
FS2/MLP and FS2/RBF    & 0.21 \\ \hline
FS1/MLP and FS2/MLP    & -0.06 \\
FS1/RBF and FS2/RBF    & -0.21 \\ \hline
\end{tabular}
\label{tab:corr}
\end{table}

There is a striking similarity between these correlation results and the
improvements obtained through combining.
When different runs of a single
classifier are combined using only one feature set, the combining improvements
are very modest. These are also the cases where the classifier correlation
coefficients are the highest. Mixing different classifiers reduces the
correlation, and in most cases, 
improves the combining results. The most drastic improvements
are obtained when two qualitatively different feature sets are used, 
which are also the cases with the lowest classifier correlations.

\subsection{Proben1 Benchmarks}
\label{sec:prob} 
In this section, examples from the Proben1 benchmark set\footnote{Available
from: {\em ftp://ftp.ira.uka.de/pub/pa\-pers/tech\-reports/1994/1994-21.ps.Z}.}
are used to study the
benefits of combining \cite{prec94}. 
Table~\ref{tab:indi} shows the test set error rate for both the MLP and 
the RBF classifiers on six different data sets taken from the Proben1 
benchmarks\footnote{These Proben1 results correspond to 
the ``pivot'' and ``no-shortcut'' architectures (A and B respectively), 
reported in~\cite{prec94}. The large error in the Proben1 no-shortcut 
architecture for the SOYBEAN1 problem is not explained.}.

\begin{table} [htb] \centering
  \caption{Performance of Individual Classifiers on the Test Set.}
  \begin{tabular}{|l||c|c||c|c||c|c||c|c||} \hline
   & \multicolumn{2}{c||}{MLP} &\multicolumn{2}{c||} {RBF} & \multicolumn{2}{c||}{Proben1-A} &\multicolumn{2}{c||} {Proben1-B} \\ \cline{2-9} 
   & Error & $\sigma$ &Error &$\sigma$ & Error & $\sigma$ & Error & $\sigma$\\ \hline \hline
   CANCER1  & 0.69 & 0.23 &  1.49 & 0.79 & 1.47  & 0.64 & 1.38 & 0.49\\
   CARD1    &13.87 & 0.76 & 13.98 & 0.95 & 13.64 & 0.85 &14.05 & 1.03\\
   DIABETES1&23.52 & 0.72 & 24.87 & 1.51 & 24.57 & 3.53 &24.10 & 1.91\\ 
   GENE1    &13.47 & 0.44 & 14.62 & 0.42 & 15.05 & 0.89 &16.67 & 3.75 \\ 
   GLASS1   &32.26 & 0.57 & 31.79 & 3.49 & 39.03 & 8.14 &32.70 & 5.34\\ 
   SOYBEAN1 & 7.35 & 0.90 &  7.88 & 0.75 & 9.06  & 0.80 &29.40 & 2.50\\ \hline
  \end{tabular}
\label{tab:indi}
\end{table}

\begin{table} [htb] \centering
\caption{Combining Results for CANCER1.}
\begin{tabular}{|l|c||c|c||c|c||c|c||c|c||} \hline
\multicolumn{2}{|c||} {Classifier(s)}& \multicolumn{2}{c||}{Ave}& \multicolumn{2}{c||}{Med} & \multicolumn{2}{c||}{Max} & \multicolumn{2}{c||}{Min}\\ \hline
{} & N & Error& $\sigma$ & Error & $\sigma$ & Error & $\sigma$ & Error & $\sigma$ \\ \hline \hline
    &3& 0.60 &0.13& 0.63 &0.17& 0.66 &0.21& 0.66 &0.21\\
MLP &5& 0.60 &0.13& 0.58 &0.00& 0.63 &0.17& 0.63 &0.17\\
    &7& 0.60 &0.13& 0.58 &0.00& 0.60 &0.13& 0.60 &0.13 \\ \hline
    &3& 1.29 &0.48& 1.12 &0.53& 1.90 &0.52& 0.95 &0.42 \\
RBF &5& 1.26 &0.47& 1.12 &0.47& 1.81 &0.58& 0.98 &0.37 \\
    &7& 1.32 &0.41& 1.18 &0.43& 1.81 &0.53& 0.89 &0.34 \\ \hline
    &3& 0.86 &0.39& 0.63 &0.18& 1.03 &0.53& 0.95 &0.42 \\
BOTH&5& 0.72 &0.25& 0.72 &0.25& 1.38 &0.43& 0.83 &0.29 \\
    &7& 0.86 &0.39& 0.58 &0.00& 1.49 &0.39& 0.83 &0.34 \\ \hline
\end{tabular}
\label{tab:cancer1}
\end{table}

\begin{table} [htb] \centering
\caption{Combining Results for CARD1.}
\begin{tabular}{|l|c||c|c||c|c||c|c||c|c||} \hline
\multicolumn{2}{|c||} {Classifier(s)}& \multicolumn{2}{c||}{Ave}& \multicolumn{2}{c||}{Med} & \multicolumn{2}{c||}{Max} & \multicolumn{2}{c||}{Min}\\ \hline
{}& N & Error& $\sigma$ & Error & $\sigma$ & Error & $\sigma$ & Error & $\sigma$ \\ \hline \hline
    &3& 13.37 &0.45& 13.61 &0.56& 13.43 &0.44& 13.40 &0.47 \\
MLP &5& 13.23 &0.36& 13.40 &0.39& 13.37 &0.45& 13.31 &0.40 \\
    &7& 13.20 &0.26& 13.29 &0.33& 13.26 &0.35& 13.20 &0.32 \\ \hline
    &3& 13.40 &0.70& 13.58 &0.76& 14.01 &0.66& 13.08 &1.05 \\
RBF &5& 13.11 &0.60& 13.29 &0.67& 13.95 &0.66& 12.88 &0.98 \\
    &7& 13.02 &0.33& 12.99 &0.33& 13.75 &0.76& 12.82 &0.67 \\ \hline
    &3& 13.75 &0.69& 13.69 &0.70& 13.49 &0.62& 13.66 &0.70 \\
BOTH&5& 13.78 &0.55& 13.66 &0.67& 13.66 &0.65& 13.75 &0.64 \\
    &7& 13.84 &0.51& 13.52 &0.58& 13.66 &0.60& 13.72 &0.70 \\ \hline
\end{tabular}
\label{tab:card1}
\end{table}

\begin{table} [htb] \centering
\caption{Combining Results for DIABETES1.}
\begin{tabular}{|l|c||c|c||c|c||c|c||c|c||} \hline
\multicolumn{2}{|c||} {Classifier(s)}& \multicolumn{2}{c||}{Ave}& \multicolumn{2}{c||}{Med} & \multicolumn{2}{c||}{Max} & \multicolumn{2}{c||}{Min}\\ \hline
{}& N & Error& $\sigma$ & Error & $\sigma$ & Error & $\sigma$ & Error & $\sigma$ \\ \hline \hline
    &3& 23.15 &0.60& 23.20 &0.53& 23.15 &0.67& 23.15 &0.67 \\
MLP &5& 23.02 &0.59& 23.13 &0.53& 22.81 &0.78& 22.76 &0.79 \\
    &7& 22.79 &0.57& 23.07 &0.52& 22.89 &0.86& 22.79 &0.88 \\ \hline
    &3& 24.69 &1.15& 24.77 &1.28& 24.82 &1.07& 24.77 &1.09 \\
RBF &5& 24.32 &0.86& 24.35 &0.72& 24.66 &0.81& 24.56 &0.90 \\
    &7& 24.22 &0.39& 24.32 &0.62& 24.79 &0.80& 24.35 &0.73 \\ \hline
    &3& 24.32 &1.14& 23.52 &0.60& 24.35 &1.21& 24.51 &1.07 \\
BOTH&5& 24.53 &0.97& 23.49 &0.59& 24.51 &1.16& 24.66 &1.02 \\
    &7& 24.43 &0.93& 23.85 &0.63& 23.85 &0.93& 24.53 &0.86 \\ \hline
\end{tabular}
\label{tab:dia1}
\end{table}

The six data sets used here are CANCER1, DIABETES1, CARD1, GENE1, GLASS1  and 
SOYBEAN1. The name and number combinations correspond to a specific 
training/validation/test set split\footnote{We are using the same
notation as in the Proben1 benchmarks.}.
In all cases, training was stopped when the test set error reached a plateau.
We report error percentages on the test set, and the standard deviation on 
those values based on $20$ runs.

CANCER1 is based on breast cancer data, obtained from the University
of Wisconsin Hospitals, from Dr. William H. Wolberg \cite{mase90,woma90}.
This set has $9$ inputs, $2$ outputs and $699$ patterns, of which $350$ is 
used for training. 
An MLP with one hidden layer of $10$ units, and an RBF network with
$8$ kernels is used with this data. 

The CARD1 data set consists of credit approval 
decisions \cite{quin87,quin92}. 
$51$ inputs are used to determine whether or not to approve the credit
card application of a customer. There are $690$ examples in this set,
and $345$ are used for training. The MLP has one hidden layer with
$20$ units, and the RBF network has $20$ kernels. 

The DIABETES1 data set is based on personal data of the Pima Indians
obtained from the National Institute of Diabetes and Digestive
and Kidney Diseases \cite{smev88}.
The binary output determines whether or not the subjects show 
signs of diabetes according to the World Health Organization.
The input consists of $8$ attributes, and there are $768$ examples in 
this set, half of which are used for training. MLPs with one 
hidden layer with $10$ units, and RBF networks with $10$ kernels
are selected for this data set.

The GENE1 is based on intron/exon boundary detection, or the detection
of splice junctions in DNA sequences \cite{noto91,tosh92}. $120$ inputs are 
used to determine whether a DNA section is a donor, an acceptor or neither.
There are $3175$ examples, of which $1588$ are used for training. The
MLP architecture consists of a single hidden layer network with $20$
hidden units. The RBF network has $10$ kernels.

The GLASS1 data set is based on the chemical analysis of glass splinters.
The $9$ inputs are used to classify $6$ different types of glass. There
are $214$ examples in this set, and $107$ of them are used for training.
MLPs with a single hidden layer of $15$ units, and RBF networks with
$20$ kernels are selected for this data set. 

The SOYBEAN1 data set consists of $19$ classes of soybean, which have
to be classified using $82$ input features \cite{mich80}. 
There are $683$ patterns in this set, of which 342 are used for training.
MLPs with one hidden layer with $40$ units, and RBF networks
with $40$ kernels are selected.

\begin{table} [htb] \centering
\caption{Combining Results for GENE1.}
\begin{tabular}{|l|c||c|c||c|c||c|c||c|c||} \hline
\multicolumn{2}{|c||} {Classifier(s)}& \multicolumn{2}{c||}{Ave}& \multicolumn{2}{c||}{Med} & \multicolumn{2}{c||}{Max} & \multicolumn{2}{c||}{Min}\\ \hline
{}& N & Error& $\sigma$ & Error & $\sigma$ & Error & $\sigma$ & Error & $\sigma$ \\ \hline \hline
    &3& 12.30 &0.42& 12.46 &0.40& 12.73 &0.55& 12.62 &0.56 \\
MLP &5& 12.23 &0.40& 12.40 &0.40& 12.67 &0.41& 12.33 &0.57 \\
    &7& 12.08 &0.23& 12.27 &0.35& 12.57 &0.31& 12.18 &0.43 \\ \hline
    &3& 14.48 &0.37& 14.52 &0.30& 14.53 &0.40& 14.42 &0.33 \\
RBF &5& 14.35 &0.33& 14.43 &0.29& 14.38 &0.24& 14.36 &0.35 \\ 
    &7& 14.33 &0.35& 14.40 &0.24& 14.28 &0.18& 14.33 &0.32 \\ \hline
    &3& 12.43 &0.48& 12.67 &0.32& 12.87 &0.65& 12.77 &0.51 \\
BOTH&5& 12.28 &0.40& 12.54 &0.35& 12.80 &0.54& 12.47 &0.65 \\
    &7& 12.17 &0.36& 12.69 &0.35& 12.70 &0.46& 12.25 &0.66 \\ \hline
\end{tabular}
\label{tab:gen1}
\end{table}

\begin{table} [htb] \centering
\caption{Combining Results for GLASS1.}
\begin{tabular}{|l|c||c|c||c|c||c|c||c|c||} \hline
\multicolumn{2}{|c||} {Classifier(s)}& \multicolumn{2}{c||}{Ave}& \multicolumn{2}{c||}{Med} & \multicolumn{2}{c||}{Max} & \multicolumn{2}{c||}{Min}\\ \hline
{}& N & Error& $\sigma$ & Error & $\sigma$ & Error & $\sigma$ & Error & $\sigma$ \\ \hline \hline
    &7& 32.07 &0.00& 32.07 &0.00& 32.07 &0.00& 32.07 &0.00 \\
MLP &5& 32.07 &0.00& 32.07 &0.00& 32.07 &0.00& 32.07 &0.00 \\
    &7& 32.07 &0.00& 32.07 &0.00& 32.07 &0.00& 32.07 &0.00 \\ \hline
    &3& 29.81 &2.28& 30.76 &2.74& 30.28 &2.02& 29.43 &2.89 \\
RBF &5& 29.25 &1.84& 30.19 &1.69& 30.85 &2.00& 28.30 &2.46 \\
    &7& 29.06 &1.51& 30.00 &1.88& 31.89 &1.78& 27.55 &1.83 \\ \hline
    &3& 30.66 &2.52& 29.06 &2.02& 33.87 &1.74& 29.91 &2.25 \\
BOTH&5& 32.36 &1.82& 28.30 &1.46& 33.68 &1.82& 29.72 &1.78 \\
    &7& 32.45 &0.96& 27.93 &1.75& 34.15 &1.68& 29.91 &1.61 \\ \hline
\end{tabular}
\label{tab:gla1}
\end{table}

\begin{table} [htb] \centering
\caption{Combining Results for SOYBEAN1.}
\begin{tabular}{|l|c||c|c||c|c||c|c||c|c||} \hline
\multicolumn{2}{|c||} {Classifier(s)}& \multicolumn{2}{c||}{Ave}& \multicolumn{2}{c||}{Med} & \multicolumn{2}{c||}{Max} & \multicolumn{2}{c||}{Min}\\ \hline
{}& N & Error& $\sigma$ & Error & $\sigma$ & Error & $\sigma$ & Error & $\sigma$ \\ \hline \hline
    &3& 7.06 &0.00& 7.09 &0.13& 7.06 &0.00& 7.85 &1.42 \\
MLP &5& 7.06 &0.00& 7.06 &0.00& 7.06 &0.00& 8.38 &1.63 \\
    &7& 7.06 &0.00& 7.06 &0.00& 7.06 &0.00& 8.88 &1.68 \\ \hline
    &3& 7.74 &0.47& 7.65 &0.42& 7.85 &0.47& 7.77 &0.44 \\
RBF &5& 7.62 &0.23& 7.68 &0.30& 7.77 &0.30& 7.65 &0.42 \\
    &7& 7.68 &0.23& 7.82 &0.33& 7.68 &0.29& 7.59 &0.45 \\ \hline
    &3& 7.18 &0.23& 7.12 &0.17& 7.56 &0.28& 7.85 &1.27 \\
BOTH&5& 7.18 &0.23& 7.12 &0.17& 7.50 &0.25& 8.06 &1.22 \\
    &7& 7.18 &0.24& 7.18 &0.23& 7.50 &0.25& 8.09 &1.05 \\ \hline
\end{tabular}
\label{tab:soy1}
\end{table}

Tables~\ref{tab:cancer1}~-~\ref{tab:soy1} show the performance of the $ave$
and $os$ combiners. 
From these results, we see that improvements are modest in general. 
However, recall that the reduction factors obtained in the previous
sections are on the {\em added} errors, not the overall error. 
For the Proben1 problems,
individual classifiers are performing well (as well or better than the results 
reported in \cite{prec94} in most cases) and it is therefore difficult to 
improve the results drastically. However, even in those cases, combining
provides an advantage: although the classification rates are not 
dramatically better, they are more reliable. Indeed, a lower standard
deviation means the results are less dependent on outside factors such as
initial conditions and training regime.
In some cases all $20$ instances of the combiner provide the same result,
and the standard deviation is reduced to zero. This can be seen in 
both the CANCER1 and SOYBEAN1 data sets. 

One important observation that emerges from these experiments is that
combining two different types of classifiers does not necessarily improve
upon (or in some cases, even reach) the error rates obtained by 
combining multiple runs of the better classifier.
This apparent inconsistency is caused by two factors.  First, as described 
in section~\ref{sec:bilin} the reduction factor is limited by the bias
reduction in most cases. If the combined bias is not lowered, the 
combiner will not outperform the better classifier. Second, as 
discussed in section~\ref{sec:comcor}, the correlation plays a major 
role in the final reduction factor. There are no guarantees that 
using different types of classifiers will reduce the correlation factors. 
Therefore, the combining of different types of classifiers, especially when 
their respective performances are significantly different (the error rate for 
the RBF network on the CANCER1 data set is over twice the error rate for 
MLPs) has to be treated with caution.

Determining which combiner (e.g. $ave$ or $med$), or 
which classifier selection (e.g. multiple MLPs or MLPs and RBFs) 
will perform best in a given situation is not generally an easy task. 
However, some information can be extracted from the experimental results.
The linear combiner, for example, appears more compatible with the MLP 
classifiers than with the RBF networks. When combining two types
of network, the {\em med} combiner often performs better than other combiners.
One reason for this is that the outputs that will be combined come from 
different sources, and selecting the largest or smallest value can favor 
one type of network over another. These results emphasize the need for
closely coupling the problem at hand with a classifier/combiner. There
does not seem to be a single type of network or combiner that can be
labeled ``best'' under all circumstances.

\section{Discussion}
\label{sec:disc}
Combining the outputs of several classifiers before making the
classification decision, has led to improved 
performance in many applications \cite{ghtu96,xukr92,zhme92}. 
This article presents a mathematical framework that underlines the reasons 
for expecting such improvements and quantifies the gains achieved. 
We show that combining classifiers in output space reduces the variance 
in boundary locations about the optimum (Bayes) boundary decision. 
Moreover, the added error regions associated with different classifiers
are directly computed and given in
terms of the boundary distribution parameters.
In the absence of classifier bias, the reduction in the added error is 
directly proportional to the reduction in the variance. 
For linear combiners, if the errors of individual classifiers are zero-mean 
i.i.d., the reduction in boundary variance is shown to be $N$, the
number of classifiers that are combined. 
When the classifiers are biased, and/or have correlated outputs, 
the reductions are less than $N$.

Order statistics combiners are discussed as an alternative to
linear methods, and are motivated by their ability to extract
the ``right'' amount of information.  
We study this family of combiners analytically, and
we present experimental results showing that 
{\em os} combiners improve upon the performance of individual classifiers.
During the derivation of the main result, the decision boundary is treated 
as a random variable without specific distribution assumptions. 
However, in order 
to obtain the table of reduction factors for the order statistics combiners, 
a specific error model needed to be adopted. Since there may be a multitude
of factors contributing to the errors, we have chosen the Gaussian model.
Reductions for several other noise models 
can be obtained from similar tables available in 
order statistics textbooks \cite{arba92,davi70}.
The expected error given in Equation~\ref{eq:err} is in
general form, and any density function can be used to
reflect changes in the distribution function. 

Although most of our analysis focuses on two classes, 
it is readily applicable to multi-class problems. In general, around
a boundary decision, the error is governed by the two (locally) dominant 
classes. 
Therefore, even in a multi-class problem, one only needs to consider the 
two classes with the highest activation values (i.e., highest posterior)
in a given localized region. 

Another important feature that arises from this
study provides a new look to the classic bias/variance dilemma. 
Combining provides
a method for decoupling the two components of the error to
a degree, allowing a reduction in the overall error. 
Bias in the individual classifiers can be reduced by using larger
classifiers than required, and the increased variance due to the larger
classifiers can be reduced during the combining stage. 
Studying the effects of this coupling between different 
errors and distinguishing situations that lead to the highest
error reduction rates are the driving motivations
behind this work. That goal is attained by clarifying the
relationship between output space combining and classification 
performance.  

Several practical issues that relate to this analysis can now
be addressed. First, let us note that since in general each individual 
classifier will have some amount of bias, the actual improvements will 
be less radical than those obtained in Section~\ref{sec:unlin}.
It is therefore important to determine how to keep the individual 
biases minimally correlated.
One method is to use classifiers with 
paradigms/architectures based on different principles. For example, 
using multi-layered perceptrons and radial basis function networks
provides both global and local information processing, 
shows less correlation than if classifiers of only one type were used.
Other methods such as resampling, cross-validation or actively
promoting diversity among classifiers can also be used, as long
as they do not adversely affect the individual classification
results.

The amount of training that is required before classifiers are combined
is also an interesting question. If a combiner can overcome
overtraining or undertraining, new training regimes could be used
for classifiers that will be combined.
We have observed that combiners do compensate for overtraining, but not
undertraining (except in cases where the undertraining is very mild). 
This corroborates well with the theoretical framework which shows
combining to be more effective at variance reduction than bias reduction.

The classification rates obtained by the order statistics combiners 
in section~\ref{sec:resu} are 
in general, comparable to those obtained by averaging.
The advantage of OS approaches should be more evident in situations where
there is substantial variability in 
the performance of individual classifiers, and the thus robust
properties of OS combining can be brought to bear upon.
Such variability in individual performance may be due to, 
for example, the classifiers being geographically distributed and
working only on locally available data of highly varying quality.
Current work by the authors indicate that this is indeed the case, 
but the issue needs to be examined in greater detail.

One final note that needs to be considered is the behavior of
combiners for a large number of classifiers ($N$). Clearly, 
the errors cannot be arbitrarily reduced by increasing $N$ indefinitely.
This observation however, does not contradict the results presented
in this analysis. For large $N$, the assumption that the errors
were i.i.d. breaks down, reducing the improvements due to each
extra classifier. 
The number of classifiers that yield the best
results depends on a number of factors, including the number of 
feature sets extracted from the data, their dimensionality, and
the selection of the network architectures.  

\nocite{wolp96a,wolp96b}

\vspace*{.1in}

\noindent
{\bf Acknowledgements:} 
This research was supported in part by
AFOSR contract F49620-93-1-0307, NSF grant ECS 9307632, and ARO 
contracts DAAH 04-94-G0417 and 04-95-10494.

\bibliographystyle{plain}

\end{document}